\documentclass[lettersize,journal]{IEEEtran}
\usepackage{amsmath,amsfonts}
\usepackage{array}
\usepackage[caption=false,font=normalsize,labelfont=sf,textfont=sf]{subfig}
\usepackage{textcomp}
\usepackage{stfloats}
\usepackage{url}
\usepackage{verbatim}
\usepackage{graphicx}
\usepackage{cite}
\usepackage{amsmath}
\usepackage{amssymb}
\usepackage{multirow}
\usepackage{xcolor}
\usepackage{colortbl}
\usepackage{arydshln}
\usepackage{comment}
\usepackage{enumitem}
\usepackage{booktabs}
\usepackage{hyperref}
\usepackage{algorithm}
\usepackage{algorithmicx}
\usepackage{algpseudocode}
\usepackage[table,x11names]{xcolor}

\begin{document}

\title{Leveraging Hierarchical Image-Text Misalignment for Universal Fake Image Detection}

\author{Daichi Zhang, Tong Zhang, Jianmin Bao, Shiming Ge,~\IEEEmembership{Senior Member,~IEEE}, \\and Sabine Süsstrunk,~\IEEEmembership{Fellow,~IEEE}
\thanks{Daichi Zhang, Tong Zhang and Sabine Süsstrunk are with the School of Computer and Communication Sciences, EPFL, 1015 Lausanne, Switzerland (e-mail: daichi.zhang@epfl.ch; tong.zhang@epfl.ch; sabine.susstrunk@epfl.ch).}
\thanks{Jianmin Bao is with Microsoft Research Asia, Beijing 100080, China (e-mail: jianbao@microsoft.com).}
\thanks{Shiming Ge is with the Institute of Information Engineering, Chinese Academy of Sciences, Beijing 100092, China, and University of Chinese Academy of Sciences, Beijing 100049, China (e-mail: geshiming@iie.ac.cn).}
\thanks{Work done when Daichi Zhang at EFPL.}
\thanks{Shiming Ge is the corresponding author (e-mail: geshiming@iie.ac.cn).}
}


\maketitle
\begin{abstract}
With the rapid development of generative models, detecting generated fake images to prevent their malicious use has become a critical issue recently. Existing methods frame this challenge as a naive binary image classification task. However, such methods focus only on visual clues, yielding trained detectors susceptible to overfitting specific image patterns and incapable of generalizing to unseen models.
In this paper, we address this issue from a multi-modal perspective and find that fake images cannot be properly aligned with corresponding captions compared to real images.
Upon this observation, we propose a simple yet effective detector termed ITEM by leveraging the \textit{i}mage-\textit{te}xt \textit{m}isalignment in a joint visual-language space as discriminative clues. Specifically, we first measure the misalignment of the images and captions in pre-trained CLIP's space, and then tune a MLP head to perform the usual detection task. 
Furthermore, we propose a hierarchical misalignment scheme that first focuses on the whole image and then each semantic object described in the caption, which can explore both global and fine-grained local semantic misalignment as clues. Extensive experiments demonstrate the superiority of our method against other state-of-the-art competitors with impressive generalization and robustness on various recent generative models.
\end{abstract}

\section{Introduction}
\label{sec_introduction}

Recent years have witnessed the rapid development of generative models, such as generative adversarial networks~(GANs)~\cite{goodfellow2014generative,karras2018progressive,karras2019style,brock2018large,park2019semantic,zhu2017unpaired} and diffusion models~\cite{dhariwal2021diffusion,nichol2021glide,rombach2022high,gu2022vector}. These generative models enable users to create high-quality synthetic images at very low cost.
However, this accessibility also presents a double-edged sword, as perpetrators can easily generate fake images for malicious use, such as using synthetic fake images to mislead the public, defame celebrities, and even fabricate evidence, leading to severe social, privacy, and security concerns~\cite{suwajanakorn2017synthesizing}. Therefore, developing general and effective fake image detectors has become a critical issue.

A common approach to tackling this issue is to frame it as a binary image classification task, discriminating between real and fake images. Typically, a dataset of real and fake images is used to train a binary classifier~\cite{wang2020cnn,Wang_2023_ICCV,tan2024rethinking}, but this approach often leads to overfitting on specific image patterns, limiting the model’s generalization capability. 
Recently, some universal detection methods~\cite{ojha2023towards,sha2023fake,liu2023forgery,zhang2025mfclip} leverage the vision encoder of Contrastive Language-Image Pre-training (CLIP)~\cite{radford2021learning} to improve the generalization of visual representations through its zero-shot abilities. However, these detectors focus solely on visual cues, neglecting the language space, which is a key driver of CLIP’s strong generalization performance. Some hybrid methods simply use text embedding by concatenation~\cite{sha2023fake} or need extra finetuning~\cite{liu2023forgery,zhang2025mfclip}, which don't fully leverage the semantic clues and still has space for improvement.
\begin{figure*}[ht]
    \centering
    \includegraphics[width=\linewidth]{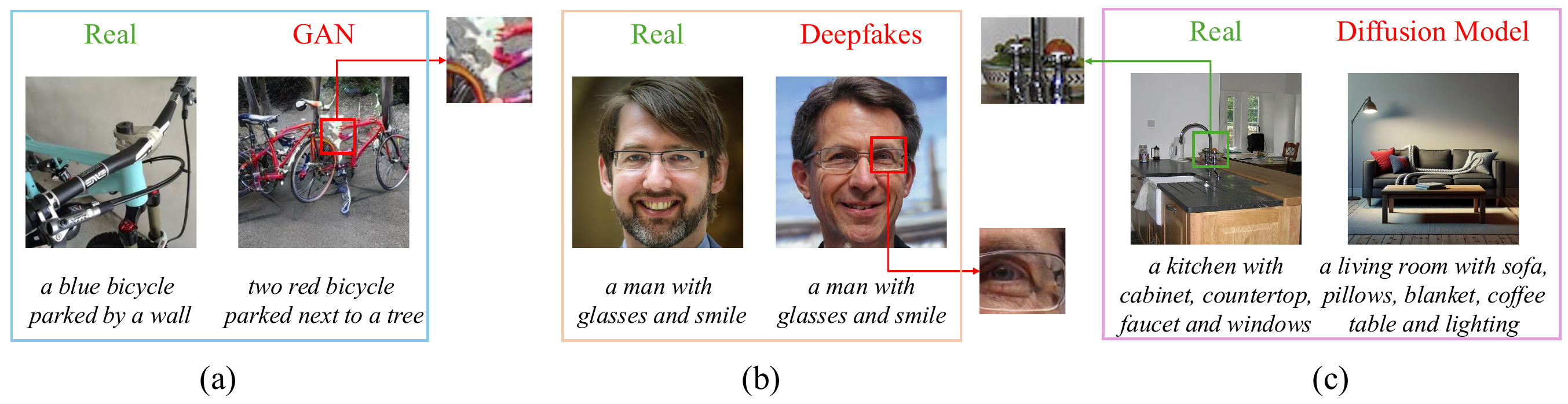}
    \vspace{-1em}
    \caption{\textbf{Motivation behind our method.} We find that the generated fake images cannot properly align with corresponding captions compared to real images, which could serve as clues for a more general and robust universal detector.}
    \label{fig:motivation}
\end{figure*}

Given these limitations, we pose the following challenge: Can we develop a universal fake image detector that generalizes to unseen generated images without relying only on visual clues? This avoids overfitting to visual-only patterns, which should lead to more general and robust detection. While most existing detectors rely solely on visual clues, we propose to tackle this challenge from a multi-modal perspective using pre-trained vision-language models (VLMs) and focus on the following question: Are the fake images properly aligned with corresponding captions as real ones?
To answer this question, we first investigate examples of real and different fake images with corresponding captions, including ProGAN~\cite{karras2018progressive} for GAN, WFIR~\cite{wfir} for deepfakes, and DALLE~\cite{dhariwal2021diffusion} for diffusion model as shown in Fig.~\ref{fig:motivation}.
We first observe that different generative models lead to different types of visual patterns, which may cause visual-only detectors cannot generalize well.
Moreover, compared to real images, some fake images, such as generated by GAN and deepfakes, exhibit local artifacts that cannot be properly described by semantic sentences, such as blurry and distortion shown in Fig.~\ref{fig:motivation} (a) and (b). These artifacts could make them misaligned with corresponding captions. 
For some other images, such as generated by text-to-image diffusion models, they are synthesized by given text prompts, which makes them highly related to the semantic information described in text, \textit{i.e.}, only contain the objects that exist in the prompt. But the captions cannot reflect all complex semantics in real scenarios, such as the hidden fruits in Fig.~\ref{fig:motivation} (c), which also makes the diffusion-generated images not properly aligned as real images.
Upon this observation, if we can incorporate the misalignment between image and text modalities as the discriminative clue for detection, we may achieve a more general and robust detector without overfitting on visual only patterns, leading to improved universal detection.

Therefore, we propose to leverage the hierarchical \textbf{i}mage-\textbf{te}xt \textbf{m}isalignment~(\textbf{ITEM}) for universal fake image detection. Specifically, we first measure the misalignment of images and their corresponding captions in the joint vision-language space of a pre-trained CLIP and then tune an MLP head for detection. Considering the detailed semantic information and artifacts in local image area, we further propose a hierarchical misalignment scheme that mines the misalignment on both the whole image and each semantic object described in the caption, which could explore both global and fine-grained local semantic clues and benefit the detection.
Our main contributions are summarized as follows:
\begin{itemize}
    \item We frame the fake image detection task from a multi-modal image-text perspective and find that the fake images cannot be properly aligned with corresponding captions compared to real images. 
    \item We propose \textbf{ITEM} to achieve universal fake image detection by leveraging the misalignment between images and captions in joint vision-language space. Moreover, a hierarchical misalignment scheme is introduced to explore both global and local fine-grained semantic misalignment.
    \item Extensive experiments on various generative models demonstrate the superiority of our proposed method against other state-of-the-art competitors with impressive generalization and robustness.
\end{itemize}
\section{Related Work}
\label{sec_related_work}

\subsection{Fake Image Detection}
With the rapid development of generative models, such as GAN~\cite{goodfellow2014generative,karras2018progressive,karras2019style,brock2018large,park2019semantic,zhu2017unpaired} and diffusion models~\cite{dhariwal2021diffusion,nichol2021glide,rombach2022high,gu2022vector}, a variety of detectors have been proposed to combat the malicious use of AI-generated fake images.
Some methods focus on the visual artifacts or traces left by generative models in fake images, 
such as the noise residual~\cite{yu2019attributing,guan2025noise}, face boundaries~\cite{li2020face}, patch-level artifacts~\cite{chai2020makes,lu2024forensicsforest}, log-perplexity~\cite{DBLP:conf/eccv/CozzolinoPNV24}, content-agnostic features~\cite{tang2025towards,tao2025sagnet}, compression traces~\cite{agarwal2017photo}, color~\cite{guo2018fake,boulkenafet2016face} and frequency clues~\cite{qian2020thinking,tan2024frequency,zeng2024toward,li2025fa}. Recent~\cite{peng2025semantic} proposes to leverage visual semantic information of human face.
Some other methods design specific representations or augmentations, such as \cite{wang2020cnn} where pre- and post-processing with data augmentation are carefully designed to build a universal GAN detector. To detect diffusion-generated images, DIRE~\cite{Wang_2023_ICCV} introduces reconstruction error based on the findings that diffusion-generated images are easier to reconstruct compared to real images. To boost generalization, recent methods have exploited pretrained models, such as UniFD~\cite{ojha2023towards} that utilizes a pre-trained CLIP-ViT~\cite{radford2021learning} model to learn the general image representation for the universal detection. NPR~\cite{tan2024rethinking} explores the artifacts left by up-sampling layers in GAN and diffusion models to serve as discriminative clues. CLIP-Flow~\cite{yuan2025clip} utilizes a normalizing flow-like unsupervised model equipped with CLIP for universal detection.

These methods, however, focus only on the difference in visual image patterns, which may lead to limited generalization on unseen generative models. Instead, we aim to address this challenge from a multi-modal perspective and focus on the image-text alignment of real and fake images. By leveraging the image-text misalignment representation on both global and local semantic levels, our method should not overfit on visual patterns and achieve a more general detection.

\begin{figure*}[ht]
    \centering
    \includegraphics[width=\linewidth]{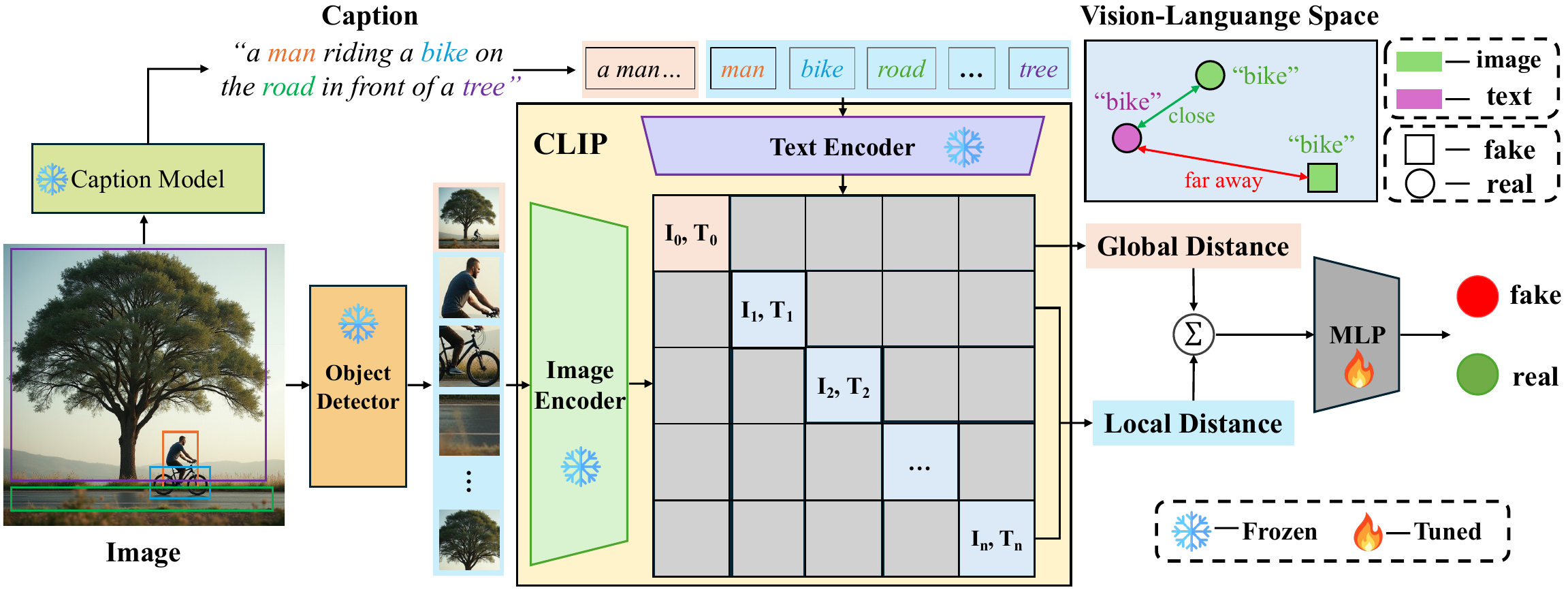}
    \vspace{-1em}
    \caption{\textbf{Overview of our proposed method.} We explore the misalignment between image and text modalities on both the global semantic clues, \textit{i.e.,} the whole image, full caption, and local fine-grained semantic clues, \textit{i.e.,} each local semantic object. After the representation learning stage, we optimize an MLP head to perform the usual fake-image detection task.}
    \label{fig:overview}
\end{figure*}

\subsection{Vision-Language Models}
Recent studies have demonstrated the great potential of vision-language models~(VLMs) in learning general visual representation and aligning visual and text concepts~\cite{liu2024visual,li2022align}. The pre-trained VLMs have been proven to have impressive transferring ability to a variety of downstream tasks~\cite{radford2021learning,singh2022flava,yuan2021florence}.
The CLIP model ~\cite{radford2021learning} could be a milestone of VLMs, as it employs transformer-based architecture~\cite{dosovitskiy2020image} with a contrastive pre-training strategy~\cite{chen2020simple} for both image and text representation learning. And there are various works following CLIP for other vision tasks~\cite{li2023blip,li2022grounded,lilanguage,xu2022groupvit}.
There are already some works~\cite{ojha2023towards,cozzolino2023raising,xu2024famsec} that use pre-trained VLMs, such as CLIP, to learn image representation for detection. 
These methods, however, use only the visual space of VLMs, which could still lead to overfitting image patterns and cause insufficient learning without fully exploring VLMs' multi-modal potential.
Whereas, we fully explore the multi-modal potential of VLMs by exploiting the misalignment between the images and generated captions in joint visual-language space at the semantic level, thus avoiding the overfitting of the visual-only image patterns and achieving improved generalization.

There are some existing methods that use the vision language models~(VLMs) for fake image detection. However, they either explored only in visual space~\cite{ojha2023towards} or simply combined cross-modal information~\cite{sha2023fake} or needed extra finetuning~\cite{liu2023forgery}. Whereas, we leverage the multi-modal misalignment between the images and captions on both global and local semantic levels, leading to a more general universal fake image detection.

\section{Methodology}
\label{sec_method}

\subsection{Image-Text Misalignment Representation}

To exploit the misalignment between image and text modalities, we first need to learn the representation of these two modalities in a given visual-language latent space. CLIP~\cite{radford2021learning} has been a milestone that optimizes an aligned vision-language space via contrastive learning. Hence, we propose to exploit the joint vision-language space of CLIP to learn the representation of image and text modality and explore their misalignment.

First, given an image $\mathbf{x}$, we employ a pre-trained caption model $\Theta_{cap}$ to generate the corresponding caption $\mathbf{p}$, which can be formulated as follows:
\begin{equation}
    \mathbf{p} = \text{Caption}(\mathbf{x}, \Theta_{cap}),
    \label{eq:cap}
\end{equation}
where the $\Theta_{cap}$ denotes the parameters of pre-trained caption model.

Then we feed the image and caption pair $(\mathbf{x}, \mathbf{p})$ into CLIP's image and text encoder, respectively, to obtain the visual and language embeddings $(\mathbf{I}, \mathbf{T})$, which can be formulated as follows:
\begin{equation}
    (\mathbf{I}, \mathbf{T}) = \text{CLIP}(\mathbf{x}, \mathbf{p}, \Theta_{clip}),
    \label{eq:clip}
\end{equation}
where the $\Theta_{clip}$ denotes the parameters of pre-trained CLIP model.

\begin{figure}[ht]
    \centering
    \includegraphics[width=\linewidth]{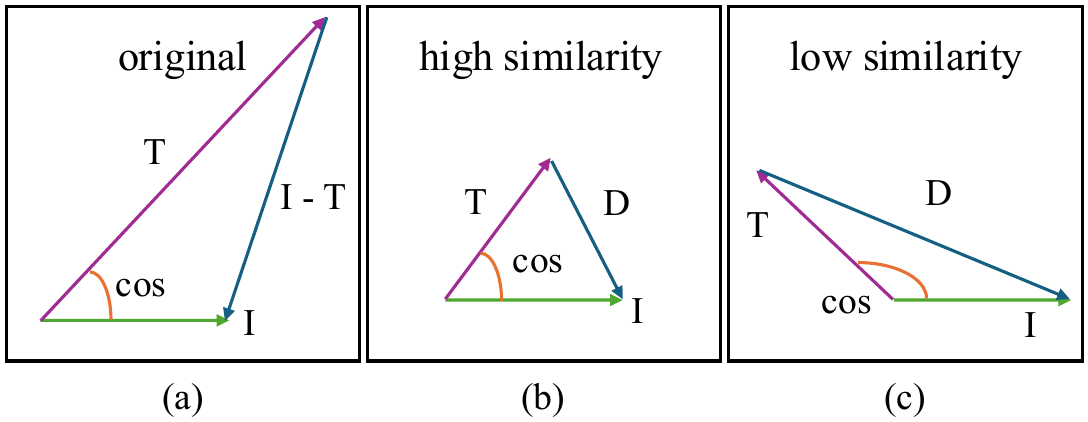}
    \vspace{-1em}
    \caption{\textbf{Image-text misalignment representation.} High cosine similarity may lead to significant distance in original CLIP space (a). Our defined misalignment representation $\mathbf{D}$ is less significant under high cosine similarity and more significant under low cosine similarity, which could properly respond to the modality misalignment.}
    \label{fig:cos_dis}
\end{figure}

Then we need to design a representation $\mathbf{D}$ to measure the misalignment of $(\mathbf{I}, \mathbf{T})$ in the joint vision-language space. As the pre-training objective of CLIP is the cosine similarity between two modalities, we propose to use the simple subtraction of the two embeddings after normalization as their distance. The reason behind this design is that the subtraction of two embeddings after normalization is related to CLIP's objective, the cosine similarity, which implies higher cosine similarity usually leads to more significant distance. Moreover, the designed representation could provide more high-dimensional information about the alignment between two embeddings from different modalities in the latent space, which should also contain informative clues for measuring cross-modal misalignment.
It is worth noting that high cosine similarity may lead to significant distance in the original CLIP space due to embedding length in different modalities, as shown in Fig.~\ref{fig:cos_dis} (a). To eliminate this effect, we normalize the embeddings before subtraction. Thus, high similarity always leads to less significant misalignment $\mathbf{D}$, while low similarity leads to more significant ones, as shown in Fig.~\ref{fig:cos_dis} (b) and (c), which is suitable and consistent with our goal. Thus, we can formulate our image-text misalignment representation as:
\begin{equation}
    \mathbf{D} = \frac{\mathbf{I}}{|\mathbf{I}|} - \frac{\mathbf{T}}{|\mathbf{T}|} ,
    \label{eq:mismatch}
\end{equation}
where $\mathbf{D}$ is our designed distance to measure the misalignment of image and text modalities.

Thus, for a given image $\mathbf{x}$, we can measure its image-text misalignment distance $\mathbf{D}$ by first using a caption model to generate its caption $\mathbf{p}$ then feeding into CLIP to obtain the embeddings and calculate. The distance serves as a clue for discriminating between fake and real images.

\subsection{Hierarchical Misalignment Scheme}

We have formulated the misalignment between a given image $\mathbf{x}$ and the corresponding caption prompt $\mathbf{p}$ in a joint CLIP latent space. The misalignment between the original image and the corresponding caption mainly focuses on the information of the whole image. This misalignment, which we term a global misalignment distance, could serve as a clue for discrimination. However, it ignores some more detailed fine-grained semantic clues that may also exist misalignment and contribute to detection, as CLIP would process all the semantic information of input image to obtain the embeddings, not particular semantic details. As shown in Fig.~\ref{fig:motivation}, the forgery artifacts in fake images usually exist in local areas. The performance could be further boosted if we could leverage more local fine-grained semantic clues. To this end, we further introduce a hierarchical misalignment scheme to explore more fine-grained local semantic forgery clues, as illustrated in Fig.~\ref{fig:overview} and described as follows.

First, for a given image $\mathbf{x}$ and its corresponding caption $\mathbf{p}$, we denote the corresponding CLIP representation as $(\mathbf{I}_0, \mathbf{T}_0)$, and we define the misalignment between the whole image and the full caption as global distance $\mathbf{D}_g$, formulated as:
\begin{equation}
    \mathbf{D}_{g} = \frac{\mathbf{I}_0}{|\mathbf{I}_0|} - \frac{\mathbf{T}_0}{|\mathbf{T}_0|},
    \label{eq:mismatch_global}
\end{equation}
where $(\mathbf{I}_0, \mathbf{T}_0) = \text{CLIP}(\mathbf{x}, \mathbf{p})$. The global distance $\mathbf{D}_{g}$ measures the cross-modal misalignment in the whole image from the global perspective.

Furthermore, to explore the fine-grained local semantic details, we focus on each semantic object existed in input image and described by the generated full caption. To this end, we introduce a pre-trained object detector $\Theta_{obj}$ and feed the input image and caption pair into it to obtain each object information, which can be formulated as follows:
\begin{equation}
    \{(\mathbf{x}_i, \mathbf{p}_i)\}_{i=1}^{i=n} = \text{ObjectDetector}(\mathbf{x}, \mathbf{p}, \Theta_{obj}),
    \label{eq:obj}
\end{equation}
where the $(\mathbf{x}_i, \mathbf{p}_i)$ pair represents each fine-grained semantic object image $\mathbf{x}_i$ after grounding and the corresponding text description, as shown in the middle of Fig.~\ref{fig:overview}.

Then, we can calculate the misalignment distance of each local semantic object following the same rule of the global distance, which can be formulated as:
\begin{equation}
    \mathbf{D}_{l}^{i} = \frac{\mathbf{I}_i}{|\mathbf{I}_i|} - \frac{\mathbf{T}_i}{|\mathbf{T}_i|} , \quad i = 1, \cdots, n
    \label{eq:mismatch_local}
\end{equation}
where $(\mathbf{I}_i, \mathbf{T}_i) = \text{CLIP}(\mathbf{x}_i, \mathbf{p}_i, \Theta_{clip})$ and the $\mathbf{D}_{l}^{i}$ is the local distance of $i$th semantic object. Then, we simply average all the local semantic objects as the final local distance:
\begin{equation}
    \mathbf{D}_{l} = \frac{1}{n} \Sigma_{i=1}^{n} \mathbf{D}_{l}^{i}, \quad i = 1, \cdots, n
    \label{eq:mismatch_local_avg}
\end{equation}
Finally, we obtain the distance that contains both global and local semantic clues by:
\begin{equation}
    \mathbf{D} = w_1 \mathbf{D}_g + w_2 \mathbf{D}_l,
    \label{eq:mismatch_final}
\end{equation}
where the $\{w_1, w_2\}$ are the hyper-parameter weights for balancing the global and local distances.

\begin{algorithm}[ht]
\caption{ITEM}
\raggedright
\textbf{Training Stage:}\\
\textbf{Input:} $N_{1}$ training image and corresponding label pairs $\{(\mathbf{x}^k, {y}^k)\}_{k=1}^{k=N_1}$, caption model $\Theta_{cap}$, clip model $\Theta_{clip}$, object detector model $\Theta_{obj}$, classifier $\Theta_{c}$\\
\textbf{Output:} trained classifier $\Theta_{c}^{*}$\\
\begin{algorithmic}[1]
\For{$k=1$ to $N_{1}$}
\State Generate caption $\mathbf{p}^k$ \Comment refer to Eq.~\ref{eq:cap}
\State Obtain embeddings $(\mathbf{I}_0^k, \mathbf{T}_0^k)$ \Comment refer to Eq.~\ref{eq:clip}
\State Compute global distance $\mathbf{D}_{g}^k$ \Comment refer to Eq.~\ref{eq:mismatch_global}
\State Detect object $\{(\mathbf{x}_i^k, \mathbf{p}_i^k)\}_{i=1}^{i=n}$ \Comment refer to Eq.~\ref{eq:obj}
\State Compute each local distance $\mathbf{D}_{l}^{(i, k)}$ \Comment refer to Eq.~\ref{eq:mismatch_local}
\State Compute final local distance $\mathbf{D}_{l}^k$ \Comment refer to Eq.~\ref{eq:mismatch_local_avg}
\State Obtain final representation $\mathbf{D}^k$ \Comment refer to Eq.~\ref{eq:mismatch_final}
\State Optimize classifier $\Theta_{c}$ with $y^k$ \Comment refer to Eq.~\ref{eq:mlp} and Eq.~\ref{eq:bce}
\EndFor
\State Obtain trained classifier $\Theta_{c}^{*}$

\end{algorithmic}
\vspace{0.3cm}
\textbf{Testing Stage:}\\
\textbf{Input:} ~~\ 
$N_{2}$ test images $\{\mathbf{x}^k\}_{k=1}^{k=N_2}$, caption model $\Theta_{cap}$, clip model $\Theta_{clip}$, object detector model $\Theta_{obj}$, trained classifier $\Theta_{c}^{*}$\\
\textbf{Output:} ~~\ 
Predicted labels $\{\hat{y}^k\}_{k=1}^{k=N_2}$\\
\begin{algorithmic}[1]
\For{$k=1$ to $N_{2}$}
\State Generate caption $\mathbf{p}^k$ \Comment refer to Eq.~\ref{eq:cap}
\State Obtain embeddings $(\mathbf{I}_0^k, \mathbf{T}_0^k)$ \Comment refer to Eq.~\ref{eq:clip}
\State Compute global distance $\mathbf{D}_{g}^k$ \Comment refer to Eq.~\ref{eq:mismatch_global}
\State Detect object $\{(\mathbf{x}_i^k, \mathbf{p}_i^k)\}_{i=1}^{i=n}$ \Comment refer to Eq.~\ref{eq:obj}
\State Compute each local distance $\mathbf{D}_{l}^{(i, k)}$ \Comment refer to Eq.~\ref{eq:mismatch_local}
\State Compute final local distance $\mathbf{D}_{l}^k$ \Comment refer to Eq.~\ref{eq:mismatch_local_avg}
\State Obtain final representation $\mathbf{D}^k$ \Comment refer to Eq.~\ref{eq:mismatch_final}
\State Predict the label $\hat{y}^k$ with trained classifier $\Theta_{c}^{*}$ \Comment refer to Eq.~\ref{eq:mlp}
\EndFor
\end{algorithmic}
\label{alg:item}
\end{algorithm}

\subsection{Fake-Image Detection Task}

After the misalignment representation learning stage, we obtain the desired distance representation that contains both global and local semantic clues. Since the $\mathbf{D}$ is still a high-dimensional vector, we tune a classification head~(a simple two-layer MLP) to perform the usual fake image detection task by predicting the label based on input distance, which can be formulated as:
\begin{equation}
    \hat{y} = \operatorname{MLP}(\mathbf{D}, \Theta_c),
    \label{eq:mlp}
\end{equation}
where $\hat{y}$ is the predicted label and ${\Theta_c}$ is the parameters of the MLP head. We employ a vanilla binary cross-entropy loss function to optimize the MLP, formulated as:
\begin{equation}
    L(y, \hat{y}) = - \Sigma_{i=1}^{N}\left( y_i \log(\hat{y}_i) + (1 - y_i) \log(1 - \hat{y}_i) \right),
    \label{eq:bce}
\end{equation}
where $N$ is the mini-batch size, $y$ is the ground-truth label, and $\hat{y}$ is the corresponding prediction of the MLP head.
We denote the trained classifier as $\Theta_{c}^{*}$, and during training, only the MLP's parameters are optimized. Then during the testing phase, we employ the same procedure on the input image to obtain its global and local misalignment distance and feed the final representation into the trained classifier $\Theta_c^{*}$ to predict the real-or-fake label.
The whole training and testing process is summarized in Algorithm~\ref{alg:item}.
\section{Experiment}
\label{sec_experiment}

\subsection{Experimental Setup}

\begin{table*}[ht]
  {\small
    \centering
    \tabcolsep=0.1cm
    \caption{\textbf{Generalization results.} Accuracy (ACC) on CNNDetection and UniformerDiffusion datasets for detecting fake images when detecting unknown generative models.}\label{tab:general_acc}
    \resizebox{1.\linewidth}{!}{
    \begin{tabular}{c ccccccc c c ccc ccc c c}
    \toprule

        \multirow{3}{*}{\shortstack[c]{Detection\\method}} & \multicolumn{7}{c}{Generative Adversarial Networks} & \multirow{2}{*}{Deepfakes} & \multicolumn{8}{c}{Diffusion Models} & Total \\
    \cmidrule(lr){2-8} \cmidrule(lr){10-17} \cmidrule(lr){18-18}

    & \multirow{2}{*}{\shortstack[c]{Pro-\\GAN}} & \multirow{2}{*}{\shortstack[c]{Cycle-\\GAN}} & \multirow{2}{*}{\shortstack[c]{Big-\\GAN}} & \multirow{2}{*}{\shortstack[c]{Style-\\GAN}} & \multirow{2}{*}{\shortstack[c]{Style-\\GAN2}} & \multirow{2}{*}{\shortstack[c]{Gau-\\GAN}} & \multirow{2}{*}{\shortstack[c]{Star-\\GAN}} & \multirow{2}{*}{\shortstack[c]{}} & \multirow{2}{*}{ADM} & \multicolumn{3}{c}{LDM} & \multicolumn{3}{c}{Glide} & \multirow{2}{*}{DALLE} & \multirow{2}{*}{Avg.}\\
    \cmidrule(lr){11-13} \cmidrule(lr){14-16}

    & & & & & & & & & & \shortstack[c]{200 steps} & \shortstack[c]{200 w/ CFG} & \shortstack[c]{100 steps} & \shortstack[c]{100 \& 27} & \shortstack[c]{50 \& 27} & \shortstack[c]{100 \& 10} & &
    
    \\ 
\midrule
\multirow{1}{*}{\shortstack[c]{ResNet-50}~\cite{he2016deep}} & {99.87} & {75.33} & {67.20} & {79.83} & {71.98} & {68.85} & {97.75} &  {64.85} & {65.75} & {66.55} & {66.70} & {67.70} & {75.65} & {79.20} & {76.55} & {55.75} & {73.72} \\

\multirow{1}{*}{\shortstack[c]{Swin-T}~\cite{liu2021swin}} & {99.77} & {91.91} & {89.04} & {83.36} & {81.55} & {88.44} & {86.14} &  {70.48} & {75.34} & {83.24} & {75.73} & {83.84} & {67.23} & {73.09} & {73.14} & {78.29} & {81.29} \\

\midrule

\multirow{1}{*}{\shortstack[c]{Patchfor}~\cite{chai2020makes}} & {92.68} & {72.90} & {65.81} & {82.11} & {81.98} & {59.13} & {88.75} &  {58.30} & {63.54} & {65.54} & {64.56} & {65.30} & {61.09} & {62.84} & {63.46} & {57.25} & {69.08} \\

\multirow{1}{*}{\shortstack[c]{F3Net}~\cite{qian2020thinking}} & {99.85} & {71.56} & {77.54} & {90.46} & {80.72} & {60.28} & {99.79} & {54.88} & {64.93} & {77.44} & {76.59} & {77.29} & {84.29} & {86.14} & {85.59} & {75.09} & {78.90} \\
\midrule

\multirow{1}{*}{\shortstack[c]{DIRE}~\cite{Wang_2023_ICCV}} & {99.83} & {67.67} & {81.75} & {84.23} & {75.73} & {80.80} & {79.40} &  {55.45} & {70.10} & {69.50} & {74.60} & {71.15} & {83.55} & {85.60} & {85.90} & {67.30} & {77.04} \\

\midrule

\multirow{1}{*}{\shortstack[c]{CNNDet}~\cite{wang2020cnn}} & {99.58} & {80.08} & {64.70} & {84.40} & {78.18} & {77.05} & {92.50} &  {78.90} & {57.25} & {54.65} & {56.35} & {55.00} & {60.55} & {64.45} & {62.15} & {56.65} & {70.15} \\

\multirow{1}{*}{\shortstack[c]{UniFD}~\cite{ojha2023towards}} & {99.65} & {93.00} & {95.70} & {85.85} & {75.55} & {99.45} & {95.30} & {81.55} & {75.20} & {94.05} & {78.45} & {94.15} & {79.65} & {81.70} & {79.25} & {86.20} & {87.17} \\

\multirow{1}{*}{\shortstack[c]{NPR}~\cite{tan2024rethinking}} & {99.90} & {77.58} & {78.90} & {93.30} & {96.43} & {75.20} & {99.60} & {64.55} & {74.70} & {81.70} & {82.40} & {82.55} & {81.45} & {83.55} & {85.45} & {63.85} & {82.57} \\

\multirow{1}{*}{\shortstack[c]{CLIP-Flow}~\cite{yuan2025clip}} & {95.76} & {91.74} & {92.72} & {75.09} & {-} & {98.26} & {90.26} & {83.16} & {60.55} & {92.45} & {80.40} & {92.80} & {82.94} & {86.75} & {84.75} & {87.30} & {86.33} \\

\multirow{1}{*}{\shortstack[c]{VIB-Net}~\cite{zhang2025towards}} & {99.99} & {97.34} & {91.05} & {71.25} & {75.30} & {98.70} & {97.85} & {83.20} & {81.55} & {97.05} & {86.70} & {97.21} & {72.89} & {75.50} & {76.63} & {94.00} & {87.26} \\

\midrule
\rowcolor{gray!30}
\textbf{ITEM} & {99.92} & {93.06} & {95.78} & {92.49} & {85.25} & {93.79} & {97.86} & {86.56} & {79.51} & {94.17} & {82.25} & {94.03} & {87.29} & {90.28} & {89.40} & {92.82} & {90.90}\\

    \bottomrule
    \end{tabular}}
    }

\end{table*}

\begin{table*}[ht]
  {\small
    \centering
    \tabcolsep=0.1cm
    \caption{\textbf{Generalization results.} Average precision (AP) on CNNDetection and UniformerDiffusion datasets when detecting unknown generative models.}\label{tab:general_ap}
    \resizebox{1.\linewidth}{!}{
    \begin{tabular}{c ccccccc c c ccc ccc c c}
    \toprule

        \multirow{3}{*}{\shortstack[c]{Detection\\method}} & \multicolumn{7}{c}{Generative Adversarial Networks} & \multirow{2}{*}{Deepfakes} & \multicolumn{8}{c}{Diffusion Models} & Total \\
    \cmidrule(lr){2-8} \cmidrule(lr){10-17} \cmidrule(lr){18-18}

    & \multirow{2}{*}{\shortstack[c]{Pro-\\GAN}} & \multirow{2}{*}{\shortstack[c]{Cycle-\\GAN}} & \multirow{2}{*}{\shortstack[c]{Big-\\GAN}} & \multirow{2}{*}{\shortstack[c]{Style-\\GAN}} & \multirow{2}{*}{\shortstack[c]{Style-\\GAN2}} & \multirow{2}{*}{\shortstack[c]{Gau-\\GAN}} & \multirow{2}{*}{\shortstack[c]{Star-\\GAN}} & \multirow{2}{*}{\shortstack[c]{}} & \multirow{2}{*}{ADM} & \multicolumn{3}{c}{LDM} & \multicolumn{3}{c}{Glide} & \multirow{2}{*}{DALLE} & \multirow{2}{*}{mAP}\\
    \cmidrule(lr){11-13} \cmidrule(lr){14-16}

    & & & & & & & & & & \shortstack[c]{200 steps} & \shortstack[c]{200 w/ CFG} & \shortstack[c]{100 steps} & \shortstack[c]{100 \& 27} & \shortstack[c]{50 \& 27} & \shortstack[c]{100 \& 10} & &
    
    \\ 
\midrule
\multirow{1}{*}{\shortstack[c]{ResNet-50}~\cite{he2016deep}} & \underline{99.99} & {83.11} & {77.42} & {98.23} & {96.23} & {78.92} & {99.88} &  {67.49} & {76.34} & {78.99} & {78.06} & {79.31} & {84.67} & {87.92} & {86.53} & {58.99} & {83.26} \\

\multirow{1}{*}{\shortstack[c]{Swin-T}~\cite{liu2021swin}} & \underline{99.99} & {99.42} & {95.80} & {92.24} & {98.34} & {96.87} & {99.76} & {76.62} & {84.99} & {92.14} & {86.55} & {92.33} & {74.70} & {80.46} & {81.53} & {87.08} & {89.93} \\

\midrule

\multirow{1}{*}{\shortstack[c]{Patchfor}~\cite{chai2020makes}} & {98.16} & {81.81} & {74.77} & {89.60} & {90.37} & {65.66} & {96.05} &  {63.37} & {71.12} & {75.49} & {74.72} & {75.33} & {69.56} & {70.56} & {71.85} & {68.32} & {77.30} \\

\multirow{1}{*}{\shortstack[c]{F3Net}~\cite{qian2020thinking}} & {99.99} & {79.20} & {89.83} & {99.03} & {99.02} & {66.86} & {100.0} & {58.16} & {75.00} & {87.92} & {84.17} & {87.46} & {92.39} & {93.89} & {93.44} & {84.99} & {86.96} \\
\midrule

\multirow{1}{*}{\shortstack[c]{DIRE}~\cite{Wang_2023_ICCV}} & {99.99} & {76.49} & {91.24} & {96.12} & {94.59} & {86.74} & {99.87} &  {53.32} & {79.74} & {77.37} & {82.59} & {79.08} & {91.69} & {93.87} & {93.85} & {74.98} & {85.72} \\

\midrule

\multirow{1}{*}{\shortstack[c]{CNNDet}~\cite{wang2020cnn}} & {99.99} & {90.77} & {87.57} & {99.26} & {98.62} & {92.70} & {98.01} &  {98.54} & {72.98} & {69.93} & {70.37} & {70.97} & {77.45} & {83.15} & {80.63} & {61.96} & {84.56} \\

\multirow{1}{*}{\shortstack[c]{UniFD}~\cite{ojha2023towards}} & {99.99} & {99.77} & {98.90} & {98.19} & {97.64} & {99.94} & {99.62} &  {96.99} & {87.00} & {97.14} & {89.11} & {96.98} & {89.69} & {91.02} & {89.65} & {93.76} & {95.34} \\

\multirow{1}{*}{\shortstack[c]{NPR}~\cite{tan2024rethinking}} & {100.0} & {97.28} & {86.93} & {98.98} & {99.42} & {78.85} & {100.0} & {61.04} & {88.31} & {89.61} & {90.03} & {90.14} & {89.02} & {90.78} & {92.01} & {71.77} & {89.01} \\

\multirow{1}{*}{\shortstack[c]{CLIP-Flow}~\cite{yuan2025clip}} & {99.86} & {98.40} & {98.38} & {89.42} & {-} & {99.84} & {98.13} & {92.23} & {76.29} & {98.49} & {94.33} & {98.46} & {95.49} & {97.07} & {96.48} & {97.15} & {95.33} \\

\multirow{1}{*}{\shortstack[c]{VIB-Net}~\cite{zhang2025towards}} & {100.0} & {99.79} & {97.10} & {93.97} & {98.12} & {99.20} & {99.78} & {95.71} & {88.70} & {86.79} & {87.68} & {89.43} & {87.32} & {88.96} & {91.47} & {95.05} & {93.69} \\

\midrule
\rowcolor{lightgray!30}
\textbf{ITEM} & {{100.0}} & {{99.88}} & {{99.38}} & {{99.27}} & {{99.19}} & {{99.95}} & {{99.96}} & {{97.93}} & {{87.27}} & {{97.87}} & {{91.56}} & {{98.06}} & {{94.75}} & {{96.58}} & {{95.57}} & {{95.59}} & {97.05}\\

    \bottomrule
    \end{tabular}}
    }

\end{table*}

\textbf{Dataset.}\label{sec:dataset}
Following recent works~\cite{wang2020cnn,ojha2023towards}, we use the images generated by following models for evaluation, including seven different GANs: (1) ProGAN~\cite{karras2018progressive}, (2) CycleGAN~\cite{zhu2017unpaired}, (3) BigGAN~\cite{brock2018large}, (4) StyleGAN~\cite{karras2019style}, (5) StyleGAN2~\cite{karras2020analyzing}, (6) GauGAN~\cite{park2019semantic}, and (7) StarGAN~\cite{choi2018stargan}, four different diffusion models with various settings: (8) ADM~\cite{dhariwal2021diffusion}, (9) LDM~\cite{rombach2022high}, (10) Glide~\cite{nichol2021glide}, (11) DALLE~\cite{ramesh2021zero}, and one high-quality (12) deepfakes method\footnote{whichfaceisreal.com}. 
To validate the performance on more recent and challenging generative models, we evaluate on recent DiffusionForensics~\cite{Wang_2023_ICCV} and GenImage~\cite{zhu2024genimage} dataset. As there exists an overlap between the two datasets, we choose ADM, Glide, and (13) VQDM~\cite{gu2022vector} from GanImage, and (14) Stable-Diffusion-v1~\cite{rombach2022high}, (15) Stable-Diffusion-v2~\cite{rombach2022high}, LDM, (16) DALLE-2, (17) Midjourney, (18) ProjGAN~\cite{sauer2021projected}, StyleGAN, (19) Diff-ProjGAN~\cite{wang2022diffusion}, and 20) Diff-StyleGAN~\cite{song2024stylegan} from DiffusionForensics dataset.
Following prior works, we train our model and other baselines on the images generated by ProGAN from~\cite{wang2020cnn}. To demonstrate our method does not highly rely on large-scale training data, we only use a subset that contains 4,0000 fake and real images, respectively.
The wide range of generative models could evaluate whether our method is generalized well for various conditions, such as classes and scenes, poor or rich semantic clues.

\noindent \textbf{Evaluation metric.}
Following prior state-of-the-art detectors~\cite{wang2020cnn,Wang_2023_ICCV,ojha2023towards}, we report accuracy~(ACC) with a fixed 0.5 threshold and an average precision~(AP) to evaluate our method and other baseline detectors. 

\noindent \textbf{Baselines.}\label{sec:baseline}
We choose the various state-of-the-art baseline detectors, including 1) ResNet-50~\cite{he2016deep}, 2) Swin-Transformer~\cite{liu2021swin}, 3) Patchforensics~\cite{chai2020makes}, 4) F3Net~\cite{qian2020thinking}, 5) DIRE~\cite{Wang_2023_ICCV}, 6) CNNDet~\cite{wang2020cnn}, 7) UniFD~\cite{ojha2023towards}, 8) NPR~\cite{tan2024rethinking}, 9) CLIP-Flow~\cite{yuan2025clip}, 10) VIB-Net~\cite{zhang2025towards}.
We categorize them into traditional image-classification backbones~(ResNet-50 and Swin-T), deepfake detectors~(Patchfor and F3Net), diffusion-generated image detectors~(DIRE), and universal detectors~(CNNDet, uniFD, NPR, CLIP-Flow, and ViB-Net).
All the above baselines are visual-only detectors, and we use the same training and testing settings for fair comparison.

\noindent \textbf{Implementation details.}
We use the pre-trained CLIP:ViT-L/14 to map the images and text prompts into 768 dimensions embeddings. The input images are center-cropped into 224 $\times$ 224, before being fed into CLIP. 
A simple fully-connected MLP is employed as our classification head, with an output dimension of 2, mapping the visual-language CLIP representation into real/fake predictions. 
To generate the caption of input images, we use BLIP-2 (blip2-opt-2.7b)~\cite{li2023blip}, and to detect each local semantic object, we use GLIP (glip-Swin-L)~\cite{li2022grounded}.
We train the classification head by 50 epochs with vanilla binary cross-entropy loss. An Adamw~\cite{loshchilov2018decoupled} optimizer with $1e-3$ learning rate and $1e-3$ weight decay is employed to optimize the training process.
We empirically set both the weights $\{ w_1, w_2\}$ of global and local distance to 1.0.
All experiments are conducted on NVIDIA A100.

\begin{table*}[ht]
  {\small
    \centering
    \tabcolsep=0.1cm
    \caption{\textbf{Generalization results on more unknown models.} Detection accuracy and average precision (ACC/AP) on more unknown diffusion models and generative adversarial networks from DiffusionForensics and GenImage datasets.}\label{tab:general_more}
    \resizebox{1.\linewidth}{!}{
    \begin{tabular}{c cccccccccccc c}
    \toprule

        \multirow{2}{*}{\shortstack[c]{Detection\\method}} & \multicolumn{8}{c}{Diffusion Models} & \multicolumn{4}{c}{Generative Adversarial Networks} & Total \\
    \cmidrule(lr){2-9} \cmidrule(lr){10-13} \cmidrule(lr){14-14}

    & \shortstack[c]{ADM} & \shortstack[c]{Glide} & \shortstack[c]{VQDM} & \shortstack[c]{SD-v1} & SD-v2 & LDM  & DALLE-2 & Mid. & \shortstack[c]{Proj-\\GAN} & \shortstack[c]{Style\\GAN} & \shortstack[c]{Diff-\\ProjGAN} & \shortstack[c]{Diff-\\StyleGAN} &  \shortstack[c]{Avg.}
    
    \\ 
\midrule
\multirow{1}{*}{\shortstack[c]{ResNet-50}~\cite{he2016deep}}& {68.00/79.95} & {71.50/83.00} & {52.35/54.89} & {76.45/79.75} & {74.65/79.91} & {58.60/84.95} &{73.47/76.41} & {90.27/83.17} & {50.40/81.19} & {55.80/93.47} & {50.65/81.13} & {93.95/94.99} & {68.01/81.07}\\

\multirow{1}{*}{\shortstack[c]{Swin-T}~\cite{liu2021swin}} & {62.68/80.77} & {58.73/74.22} & {66.18/81.55} & {53.48/61.24} & {64.47/73.94} & {81.69/94.96} & {76.18/77.61} & {90.79/80.03} & {50.88/71.29} & {69.78/86.71} & {50.23/55.72} & {87.79/91.69} & {67.74/77.48}\\

\midrule

\multirow{1}{*}{\shortstack[c]{Patchfor}~\cite{chai2020makes}} & {56.96/63.91} & {58.98/65.57} & {64.24/75.47} & {73.14/89.44} & {76.14/88.66} & {81.38/92.71} & {82.65/94.95} & {91.56}/{97.97} & {63.86/80.11} & {64.57/80.10} & {64.54/79.85} & {84.89/94.60} & {71.91/83.61}\\

\multirow{1}{*}{\shortstack[c]{F3Net}~\cite{qian2020thinking}} & {72.29/81.20} & {73.39/82.53} & {66.13/76.12} & {78.14/93.37} & {80.19/89.11} & {87.89/97.40} & {90.79/96.81} & {87.29/95.28} & {72.49/96.97} & {88.10/95.35} & {65.98/95.71} & {92.49/99.25} & {79.60/91.59}\\

\midrule

\multirow{1}{*}{\shortstack[c]{DIRE}~\cite{Wang_2023_ICCV}} & {75.25/85.47} & {81.45/90.49} & {66.85/76.79} & {74.05/81.80} & {73.10/88.97} & {80.65/98.48} & {73.87/94.63} & {63.91/86.80} & {61.40/51.16} & {75.60/88.25} & {61.40/55.44} & {79.85/94.59} & {72.28/82.74}\\

\midrule

\multirow{1}{*}{\shortstack[c]{CNNDet}~\cite{wang2020cnn}} & {56.45/72.39} & {57.90/74.36} & {54.20/61.73} & {50.25/74.85} & {50.05/64.16} & {53.65/78.37} & {66.80/63.15} & {90.82/91.42} & {56.00/88.07} & {82.15/98.65} & {55.25/85.57} & {95.35/99.87} & {64.07/79.38}\\

\multirow{1}{*}{\shortstack[c]{UniFD}~\cite{ojha2023towards}} & {73.15/85.62} & {61.95/72.57} & {84.30/93.07} & {74.20/93.91} & {65.25/90.48} & {85.00/90.19} & {96.50/98.71} & {73.00/95.44} & {88.80/97.61} & {80.70/96.49} & {87.90/94.98} & {83.85/97.35} & {79.55/92.20}\\

\multirow{1}{*}{\shortstack[c]{NPR}~\cite{tan2024rethinking}} & {70.80/81.72} & {71.95/88.82} & {67.80/71.91} & {81.25/88.85} & {76.40/88.95} & {80.45/84.97} & {66.67/71.32} & {90.91/87.25} & {79.45/97.36} & {85.95/96.18} & {84.65/99.02} & {95.75/99.12} & {79.34/87.96}\\

\midrule
\rowcolor{lightgray!30}
\textbf{ITEM} & {87.24/93.00} & {79.11/90.59} & {86.93/93.87} & {80.52/94.11} & {77.86/89.90} & {89.79/97.58} &  {91.19/98.82} & {89.75/98.08} & {92.29/97.99} & {89.01/96.55} & {88.13/96.03} & {95.52/99.34} & {87.29/95.49}\\

    \bottomrule
    \end{tabular}}
    }

\end{table*}

\subsection{Comparison to State-of-the-Art}

\textbf{Generalization to unknown models.}
We begin by evaluating the detectors' generalization to unknown generative models, which is a critical challenge in this field. First, we evaluate them on the CNNDet~\cite{wang2020cnn} and UniformerDiffusion~\cite{ojha2023towards} datasets, and the ACC/AP results are shown in Tab.~\ref{tab:general_acc}\&\ref{tab:general_ap}.
From the results, we observe that naive detectors, such as ResNet-50 and Swin-T, cannot achieve the desired performance on the unknown generative models. The detector designed for CNN-generated images, such as CNNDet, suffers from detecting diffusion-generated images, and the same as diffusion-generated image detectors, such as DIRE. The universal detectors, including UniFD and NPR, achieve considerable performance on various unseen models. But, they still encounter performance drops on unseen models, such as StyleGAN2 for UniFD and DALLE for NPR, which we assume is caused by the unseen model architectures or distributions. 
Whereas, our proposed method achieves impressive performance on various kinds of generative models, with an average improvement of 11.33\% AP compared to DIRE, 1.71\% compared to UniFD, and 8.04\% compared to NPR. 
Note that the UniFD detector relies on only the visual space of CLIP, which leads to a performance drop, \textit{i.e.} 8.79\% ACC and 5.51\% AP drops on Glide. This provides evidence for our assumption that leveraging multi-modal clues instead of focusing only on visual space benefits the detection.

Furthermore, we conduct experiments on more unseen models from recent DiffusionForensics and GenImage datasets as shown in Tab.~\ref{tab:general_more}. The results demonstrate that our proposed universal detector is general to various unseen models.

\begin{figure*}[ht]
    \centering
    \includegraphics[width=\linewidth]{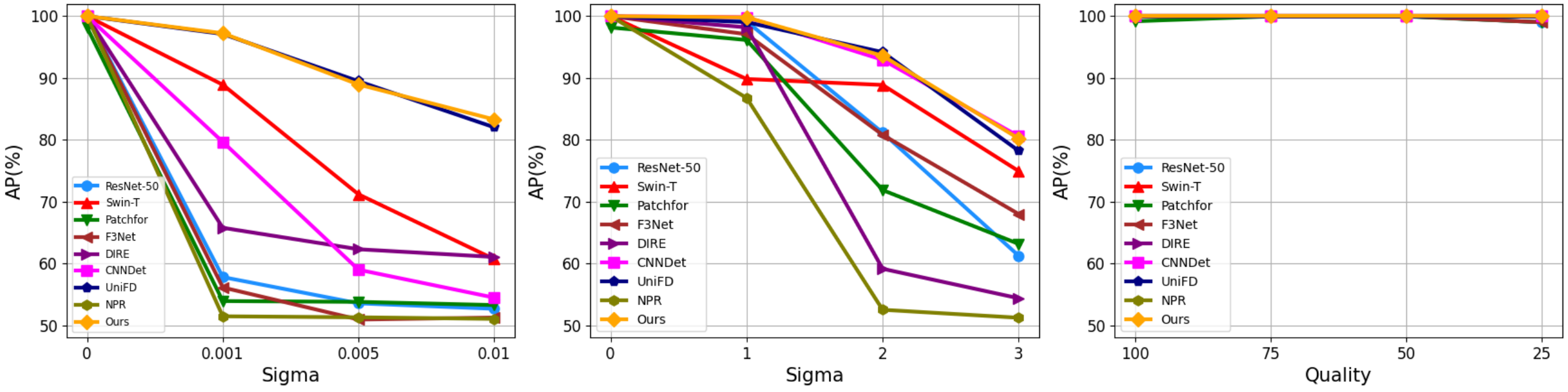}
    \vspace{-1em}
    \caption{\textbf{Robustness results to unseen perturbations.} Average precision (AP) under three different types of perturbations with three different severity levels: Gaussian Noise~($\sigma = 0.001,0.005,0.01$), Gaussian Blur~($\sigma = 1,2,3$), and JPEG Compression~($quality = 75,50,25$)~(from left to right).}
    \label{fig:robust}
\end{figure*}

\noindent \textbf{Robustness to unseen perturbations.}\label{sec:robust}
The robustness is also a critical concern for current detectors, as there are various post-preprocessing perturbations in real scenarios, such as compression. We evaluate detectors' robustness against three common types of perturbations on images generated from ProGAN~(the same as the training set), including Gaussian Noise, Gaussian Blur, and JPEG Compression following~\cite{wang2020cnn,Wang_2023_ICCV}. For each perturbation, we consider three different severity levels: $\sigma = 0.001,0.005,0.01$ for Gaussian Noise, $\sigma = 1,2,3$ for Gaussian Blur, and $quality = 75, 50, 25$ for JPEG Compression. The results are shown in Fig.~\ref{fig:robust}.
We observe that our method suffers less from the three different perturbation types, with only a very slight performance drop compared to other baselines, especially under Gaussian Noise and Gaussian Blur.
This indicates that leveraging the visual-language misalignment in a pre-trained CLIP model leads to a more robust representation compared to using only visual image patterns.

\begin{figure*}[ht]
    \centering
    \includegraphics[width=\linewidth]{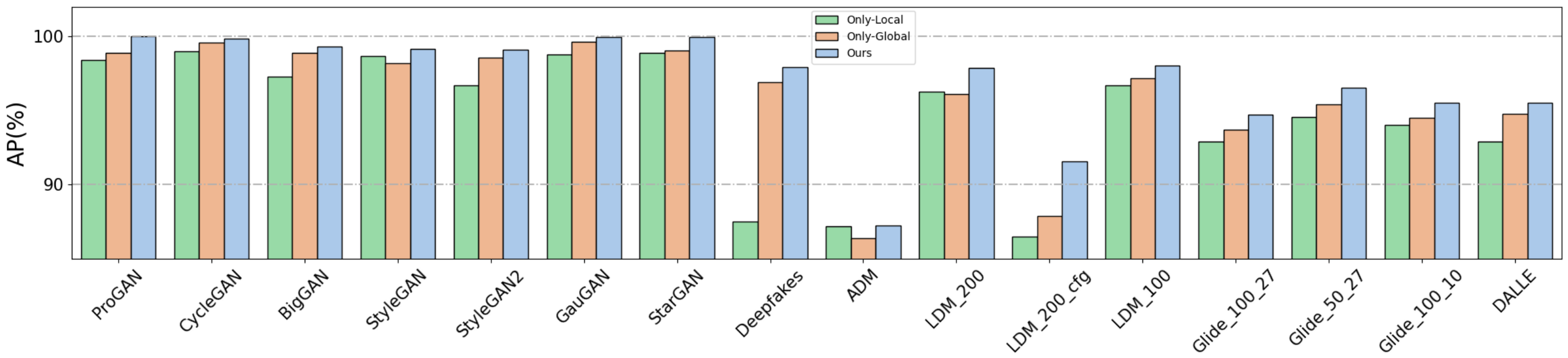}
    \vspace{-1em}
    \caption{\textbf{Ablation study on different distances.} The average precision~(AP(\%)) is reported. We observe that our proposed local and global distances could both achieve impressive performance, and the performance is further boosted with equipped both.}
    \label{fig:ablation_distance}
\end{figure*}

\begin{figure*}[ht]
    \centering
    \includegraphics[width=\linewidth]{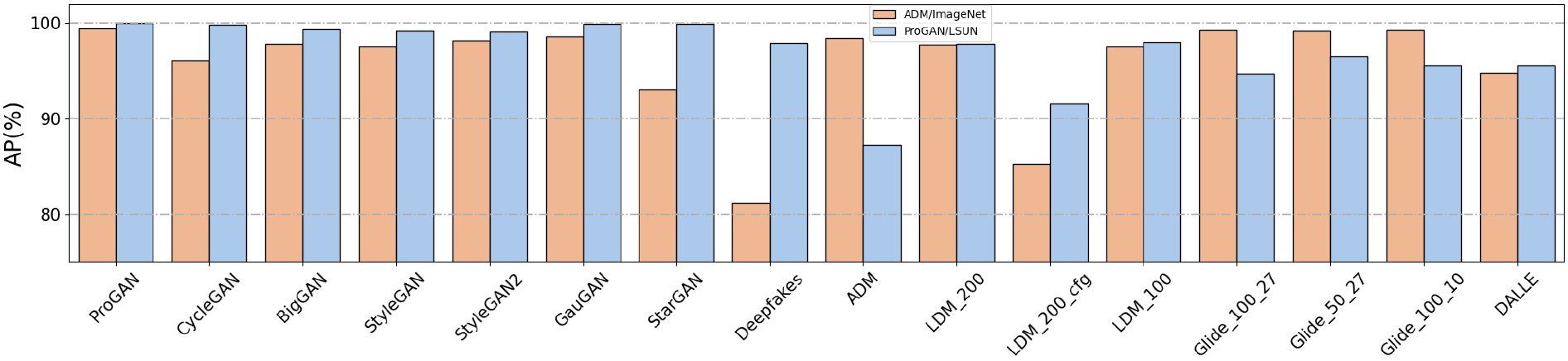}
    \vspace{-1em}
    \caption{\textbf{Ablation study on different training datasets.} The average precision~(AP(\%)) is reported. Our proposed ITEM achieves impressive results, regardless of the generative models (e.g., GAN or diffusion models) and image source (e.g., ImageNet or LSUN).}
    \label{fig:ablation_dataset}
\end{figure*}

\begin{figure*}[ht]
    \centering
    \includegraphics[width=\linewidth]{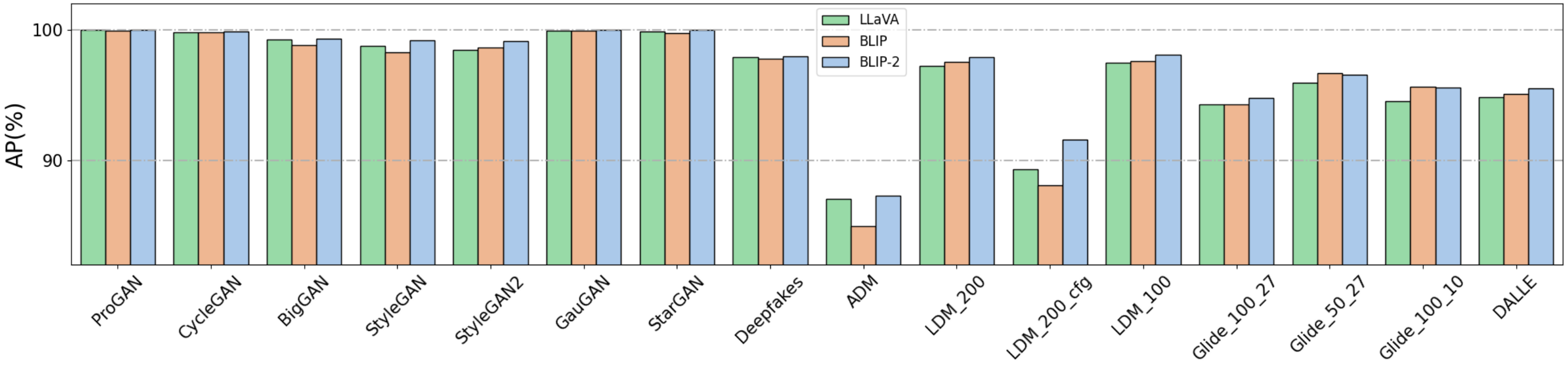}
    \vspace{-1em}
    \caption{\textbf{Ablation study on different caption models.} Our method is robust to different caption models, which indicates that the misalignment between image and caption is a general phenomenon.}
    \label{fig:ablation_caption}
\end{figure*}

\subsection{Ablation Study}

\noindent \textbf{Effect of different distances.}
To demonstrate that both our designed global and local distances contribute to the improved performance of the detection, we conduct an ablation study by employing the following variants: (i) only-local distance, (ii) only-global distance, and (iii) both distances. The results are shown in Fig~\ref{fig:ablation_distance}. We observe that using only the local or global distance could both achieve an impressive performance, and the performance is further boosted when both are employed. 
The results support our hypothesis that the misalignment exists in both the whole image and local areas. Exploring both global and more detailed fine-grained misalignment clues leads to further improvements.
Besides, we observe that the performance is still impressive when only using global distance. This provides a flexible choice for efficiency: when the object detector is not available or too time-consuming, we can still get satisfying results by only using the global distance.

\noindent \textbf{Effect of different training datasets.}
To evaluate whether our detector is universal when training data changes, we conduct experiments by using different generative models and image sources as the training set. We consider both the GAN and diffusion models and evaluate the following two variants: (i) ADM~\cite{dhariwal2021diffusion} trained on ImageNet~\cite{russakovsky2015imagenet}, and (ii) ProGAN~\cite{karras2018progressive} trained on LSUN. 
The results are shown in Fig.~\ref{fig:ablation_dataset}. 
We observe that our method, when trained on diffusion-generated images can achieve impressive performance on GANs, and the same for detecting diffusion-generated images, when trained on GAN-generated images.
Additionally, different training datasets can achieve similar impressive performance, irrespective of the generative models or image sources. This provides more evidence that our motivation and proposed detector are general and universal to different unseen generative models, irrespective of different training datasets, \textit{i.e.}, generative models or image sources.

\begin{figure*}[ht]
    \centering
    \includegraphics[width=\linewidth]{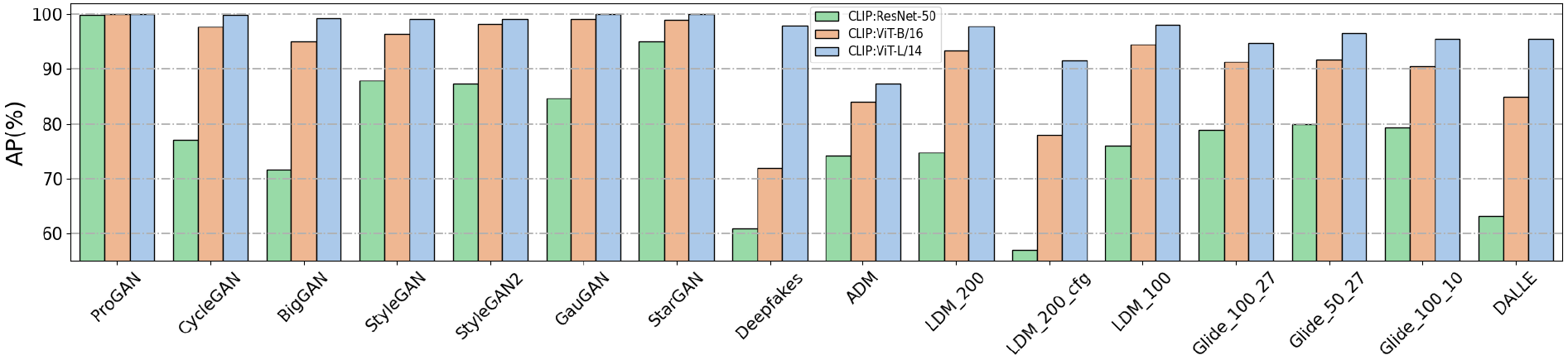}
    \vspace{-1em}
    \caption{\textbf{Ablation study on different CLIP architectures.} The results indicate that the detection performance benefits from a larger CLIP backbone architecture.}
    \label{fig:ablation_clip}
\end{figure*}

\begin{figure}[ht]
    \centering
    \includegraphics[width=\linewidth]{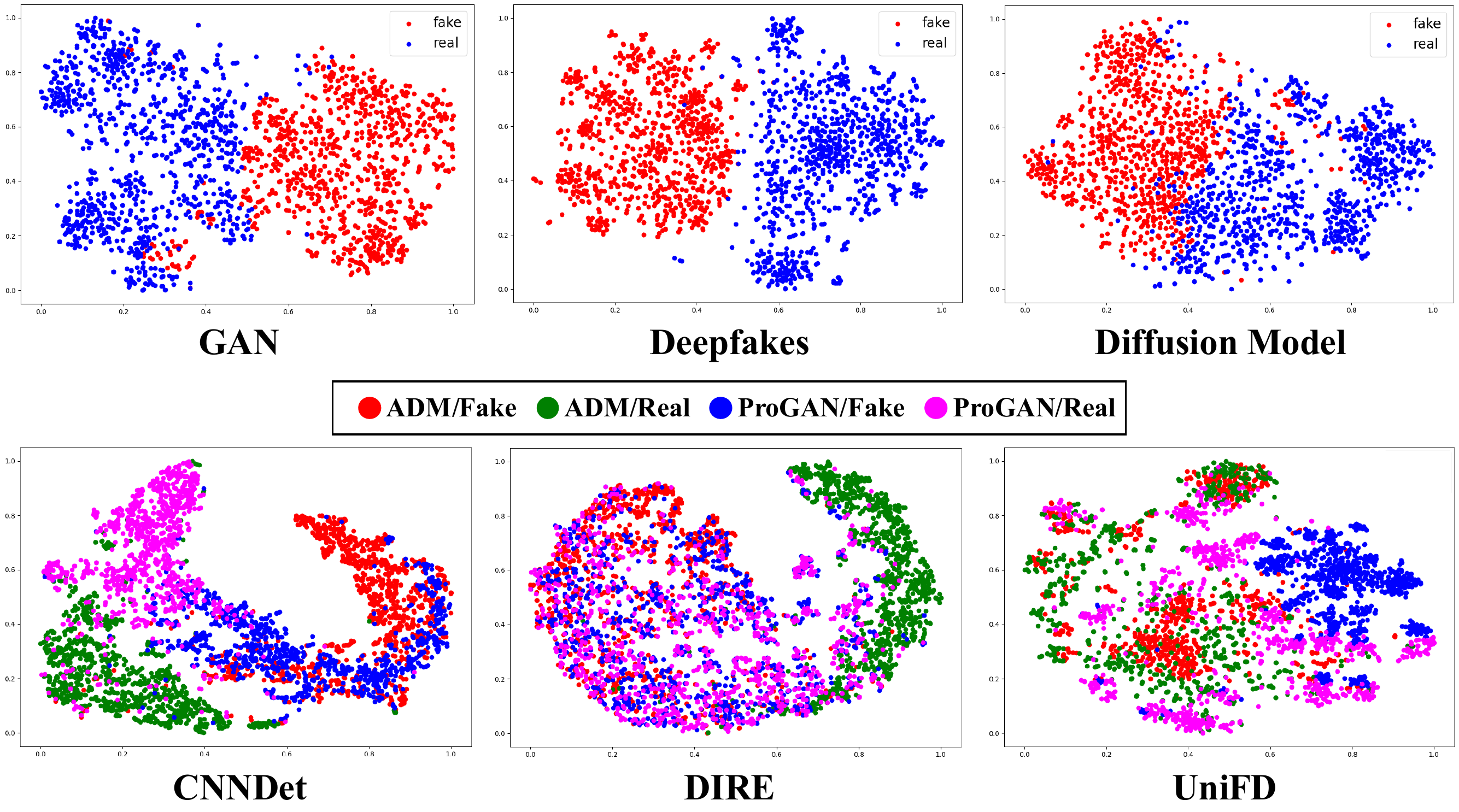}
    \vspace{-1em}
    \caption{\textbf{The t-SNE visualization of our representation and others}, which indicates our representation preserves strong discriminability in distinguishing fake images from real ones.}
    \label{fig:tsne}
\end{figure}

\noindent \textbf{Effect of different caption models.}
We conduct further ablations to demonstrate our method's effectiveness when employing different caption models. Specifically, we consider following different caption models: (i) LLaVA~\cite{liu2024visual}, (ii) BLIP~\cite{li2022blip}, and (iii) BLIP-2~\cite{li2023blip}. Note that the prompt we use for LLaVA is: "Please generate a one-sentence caption for the input image." The results are shown in Fig.~\ref{fig:ablation_caption} and we observe that our method still achieves impressive performance when employing a different caption model, with only a slight drop compared to BLIP-2. This indicates that our method is general and applicable to different caption models, and the performance is not overfit to certain caption patterns. This evidence also demonstrates that the misalignment between image and text in generated fake images generally exists across different caption models.

\noindent \textbf{Effect of different CLIP architectures.}
We conduct experiments to investigate the effect of different CLIP backbone architectures, including: (i) CLIP:ResNet-50, (ii) CLIP:ViT-B/16, and (iii) CLIP:ViT-L/14. The results are shown in Fig.~\ref{fig:ablation_clip}. 
From the results, we observe that variations in CLIP could influence the performance. 
Specifically, the transformer-based CLIP architecture performs better than ResNet-50, which could be explained by its large-scale architecture and the long-range receptive field introduced by the attention blocks. ViT-L/14 also achieves higher performance than ViT-B/16, which could also be attributed to the larger backbone. This implies that a suitable pre-trained vision-language space benefits the detection performance, such as transformer-based CLIP.

\subsection{Visualization}
To analyze whether our method could effectively distinguish the real and fake images, we visualize the distance representation by using t-SNE~\cite{tSNE} on different models, including ProGAN~\cite{karras2018progressive} for GAN model, LDM~\cite{rombach2022high} for diffusion model, and StarGAN~\cite{choi2018stargan} for deepfakes. 
The results are presented on top of Fig.~\ref{fig:tsne}, which shows that the representations of real and fake images are clustered with a clear discrepancy margin in latent space for all three different generative models. This indicates that our representation has strong discriminability and generalizability in distinguishing real and fake images. 
Besides, we also visualize the representation of other three baselines: CNNDet~\cite{wang2020cnn}, DIRE~\cite{Wang_2023_ICCV}, and UniFD~\cite{ojha2023towards} on GAN and Diffusion models. The results are shown at the bottom of Fig.~\ref{fig:tsne}. From the results, we observe that these competitors can not distinguish real and fake properly: the real and fake images are clustered closely.
\section{Conclusion}
\label{sec_conclusion}

In this paper, we find that fake images cannot be properly aligned with corresponding captions compared to real images.
Upon this observation, we reframe the fake image detection from a multi-modal image-text perspective and propose the \textbf{ITEM} to achieve universal fake image detection.
Furthermore, we introduce a hierarchical misalignment scheme to mine both global and fine-grained local semantic misalignment as clues. 
Extensive experiments demonstrate our method's superiority against other state-of-the-art competitors with impressive generalization and robustness. 
We hope our method can provide insight on how to formulate the AI-generated image detection task from a multi-modal perspective and how to fully leverage large pre-trained models to detect AI-generated content~(AIGC) for future research.




\bibliographystyle{IEEEtran}
\bibliography{IEEEabrv, main}



\vfill

\end{document}